%% file: acl_latex.tex
\documentclass[11pt]{article}

\usepackage[preprint]{acl}

\usepackage{times}
\usepackage{latexsym}
\usepackage[T1]{fontenc}
\usepackage[utf8]{inputenc}
\usepackage{microtype}
\usepackage{inconsolata}

\usepackage{amsmath,amssymb,amsfonts}
\usepackage{graphicx}
\usepackage{booktabs}
\usepackage{tabularx,array}
\usepackage{textcomp}
\usepackage{xcolor}
\usepackage{ragged2e}
\newcolumntype{L}[1]{>{\RaggedRight\arraybackslash}p{#1}}
\title{Jev for Scientific Decisions: Evaluating Semantic Choices and Their Consequences}

\author{
  Boyuan Deng \\
  Johns Hopkins University \\
  Baltimore, USA \\\And
  Shuyi Fan \\
  Columbia University \\
  New York, USA \\\AND
  Hongyang Zhang \\
  The Hong Kong Polytechnic University \\
  Hong Kong, China \\\And
  Xinhong Xie \\
  The Pennsylvania State University \\
  University Park, USA \\\AND \normalfont\small \textbf{Correspondence:} \href{mailto:bdeng8@jh.edu}{bdeng8@jh.edu}
}

\begin{document}
\maketitle

\begin{abstract}
    Scientific workflows often require choosing among known relations before a deterministic calculation can proceed. Whether observations share a culture, treatment or reference standard can change the scientific meaning of the resulting count or comparison. Once the evidence is prepared and the calculation is written as code, this decision reduces to selecting one of the listed candidates. We evaluate Jev on this step, using a harness that follows its documented guidance. The study compares twelve model configurations on ten scientific cases built from published sources. Each case contains two Choices, which are questions with listed candidates. Each case is repeated five times per configuration. We measure selections, downstream outputs and final labels separately. Jev tied with five other configurations at complete semantic correctness and achieved the lowest median latency and mean cost. On one Choice, whether samples came from one culture, three comparison configurations made seven wrong selections that changed downstream counts while preserving the correct final label. These results support using Jev to select relations when the evidence and candidates are prepared in advance. They also show that evaluating it requires checking the selections and downstream outputs that a workflow will reuse.
\end{abstract}

\section{Introduction}
\label{sec:introduction}

Scientific workflows often turn on small decisions about what an observation represents. Ten measurements may describe independent cultures or repeated samples from one culture. That distinction determines what a program should count and which claims the evidence supports. Similar decisions arise when identifying shared calibration standards, paired simulations or experimental assignments. Such decisions have a short list of possible answers. Once the candidates are listed and the calculation is written as code, what remains for the model is one selection that depends on reading scientific language correctly.

Jev is a model that evaluates typed questions against supplied data and returns structured results without generating text \citep{typesafe2026introduction}. Its Choice interface selects from explicit candidates \citep{typesafe2026choice}. The documented division of labor assigns numerical operations to code and semantic judgments to the model \citep{typesafe2026jaggedness}. We ask whether a model makes these semantic judgments correctly at low cost and latency, whether doing so requires reasoning tokens, and whether evaluation captures their downstream consequences.

Existing benchmarks evaluate answers to scientific questions and labels of scientific claims. PubMedQA pairs research questions with abstract context \citep{jin2019pubmedqa}. The Jevals benchmark uses a binary version to evaluate Jev \citep{jevals2026methodology}. SciFact evaluates scientific claims together with supporting evidence \citep{wadden2020fact}. More recently, \citet{ho2025table} showed that correct scientific claim labels can coexist with incorrect cell-level rationales. \citet{sprague2025cot} found that chain-of-thought prompting helps mainly on mathematical and symbolic tasks. PAL \citep{gao2023pal} delegates such steps to an interpreter. Our harness assigns these operations to fixed code. We examine how selections determine subsequent calculations and whether correct final labels are accompanied by correct downstream outputs.

We contribute cases built from published sources, a shared harness and a comparison of Jev with eleven other configurations. Our analysis connects semantic correctness to downstream outputs and final labels. It also compares latency, cost and reported reasoning tokens.

\section{Scientific Decisions and Evaluation}
\label{sec:evaluation}

\subsection{Cases and the shared harness}
\label{sec:cases}

The model comparison uses twenty scientific Choices in ten cases, with two Choices per case. A Choice is one question with a fixed list of candidates. The Choices ask what counts as one experimental unit, whether units share a history such as a common culture, whether observations depend on one another, and which protocol applies to which observations. Each case also has a claim that we constructed about the source study. The claim consists of clauses. Shared code judges each clause as supported, contradicted or unresolved from the selections and downstream outputs. A clause is unresolved when these neither support nor contradict it. Sources span manufacturing and materials, physical simulations, climate, metrology and biological experiments. These cases are ten of the twenty case groups in a collection of forty scientific Choices. We call the twenty Choices answered by the models \emph{model-resolved} and the other twenty \emph{code-resolved}. Deterministic parsers written for the prepared passages answer the code-resolved Choices without a model call. A parser leaves its Choice unresolved unless the passage matches the expected wording in full. Appendix \ref{app:cases} lists the cases and their reference interpretations.

Each configuration receives the same prepared evidence passages, structured records, question instructions and candidate relations. These inputs adapt or transcribe the source documents. Reference answers and the constructed claims are withheld from all configurations.

Following Jev's guidance, the harness asks atomic questions, points each question to the supplied evidence and keeps counting, filtering, arithmetic and label composition in code. Jev uses its native Choice interface. The other configurations return candidate identifiers through a constrained JSON schema. Both Choices of a case are submitted in one request.

\begin{figure*}[t]
  \centering
  \includegraphics[width=0.95\textwidth]{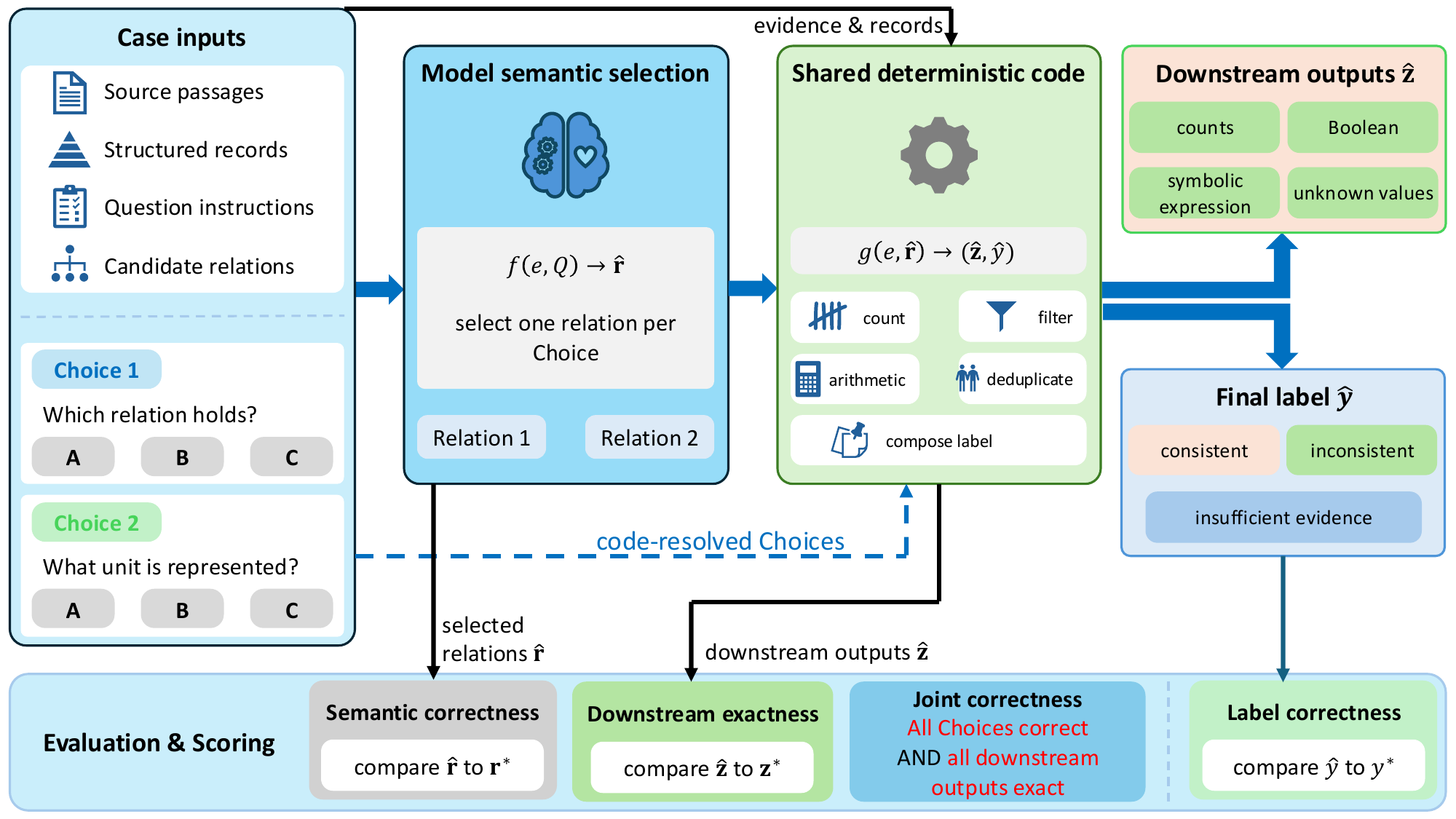}
  \caption{The evaluation workflow, from case inputs to scoring. Code-resolved Choices bypass the model.}
  \label{fig:workflow}
\end{figure*}

\subsection{Three levels of correctness}
\label{sec:metrics}

Let $e$ denote the supplied evidence and records, and $Q$ the Choices with their candidates. The model $f$ selects one candidate for each Choice. These selections form $\hat{\mathbf r}$. The shared deterministic program $g$ applies the case's computation and label rules to produce downstream outputs $\hat{\mathbf z}$ and a final label $\hat y$:
\begin{equation}
 \hat{\mathbf r}=f(e,Q),\qquad
 (\hat{\mathbf z},\hat y)=g(e,\hat{\mathbf r}).
 \label{eq:workflow}
\end{equation}
Figure~\ref{fig:workflow} shows this division. The reference answers, downstream outputs and label are $\mathbf r^*$, $\mathbf z^*$ and $y^*$.

\textbf{Semantic correctness} requires the selection for Choice $k$ to match its reference answer exactly: $\hat r_k=r_k^*$. \textbf{Downstream exactness} requires agreement on every output field that the case declares (Appendix Table~\ref{tab:output-fields}). These fields form the downstream outputs and include counts, Booleans, symbolic expressions and unknown values. They exclude clause judgments and labels. \textbf{Label correctness} requires $\hat y=y^*$. The program $g$ combines the clause judgments into the final label. Any contradicted clause makes it \texttt{inconsistent}. Otherwise, any unresolved clause makes it \texttt{insufficient\_evidence}. A claim whose clauses are all supported is \texttt{consistent}. We also report \textbf{joint correctness}, requiring all selections and downstream outputs to agree. Appendix~\ref{app:metrics} specifies the output fields and scoring procedure.

The program in Equation~\ref{eq:workflow} can map different selections to the same downstream outputs or final label. It can also compute exactly from a wrong selection and return wrong downstream outputs. The reference label of all ten model-resolved cases is \texttt{inconsistent}, so an always-\texttt{inconsistent} baseline solves the label task. We therefore compare configurations on semantic and joint correctness.

\subsection{Experimental comparison}
\label{sec:setup}

We compared twelve configurations, each a model with fixed service settings, on all ten cases, each repeated five times. Each configuration therefore had fifty planned requests and one hundred planned selections. Configurations were evaluated in three batches, with serial requests, shuffled case order and rotating configuration order within each batch. Repetitions measure variation on the same inputs. Two checks require no model calls. A code-only run confirms that the parsers reproduce the reference answers of all twenty code-resolved Choices. A reference-oracle run passes the reference answers for all forty Choices to the shared code, confirming that the derivation code reproduces every reference downstream output and reference label.

A request is successful when the response arrives and gives a valid candidate for both Choices. Any other request, including one with an invalid candidate identifier, counts as failed. The primary scores divide by all planned requests or selections, so a failed request earns no credit. We also report correctness conditional on a successful request (Appendix~\ref{app:observed}). Each request was sent once, with no fallback to another hosting provider. We requested medium reasoning effort for the OpenAI and Qwen configurations. DeepSeek and Kimi offer no separate medium setting, so we requested high effort for them. Anthropic configurations used adaptive thinking, in which the model decides whether and how much to reason, at medium reasoning effort. The eleven comparison configurations had an output limit of 32,768 tokens. Jev requests set neither reasoning effort nor an output limit. Exact model identifiers and service settings are provided with the released code at \url{https://github.com/bydeng01/scientific-decision-eval}.

Latency is the time from sending a request to receiving the full response, measured at the client. Cost is the charge reported by the provider, or an estimate from token usage and the applicable prices when no charge is reported. All latency and cost statistics are computed over successful requests. We report the median and nearest-rank 95th percentile (p95) of latency and the mean cost.

\begin{table*}[t]
\centering
\small
\begin{tabular*}{\textwidth}{@{\extracolsep{\fill}}lrrrrrrrr@{}}
\toprule
 & \multicolumn{4}{c}{Requests and correctness} & \multicolumn{4}{c}{Latency and cost (successful requests)} \\
\cmidrule(lr){2-5}\cmidrule(l){6-9}
Configuration & Successful & Semantic & Downstream & Label & Reasoning & Median & p95 & Cost \\
 & /50 & /100 & $=$ Joint /50 & /50 & tokens & (s) & (s) & (\$/1,000) \\
\midrule
\csname @@input\endcsname ./main_table_rows.tex
\bottomrule
\end{tabular*}
\caption{Comparison on twenty Choices in ten cases, repeated five times per configuration. Correctness is computed over all planned requests. Downstream and joint counts coincide because every wrong selection changed an output field. Cost is the mean cost per 1,000 requests. The marker \textsuperscript{a} denotes reported charges and \textsuperscript{b} usage-based estimates. Reasoning tokens are medians. Jev does not report them.}
\label{tab:main}
\end{table*}

\section{Results}
\label{sec:results}

\subsection{Semantic and downstream correctness}

Jev answered all one hundred planned Choices correctly, and its downstream outputs were exact in all fifty requests (Table~\ref{tab:main}). GPT-5.6 Sol \citep{openai2026gpt56}, GPT-6 Astra \citep{openai2026gpt6astra}, Claude Sonnet 5 \citep{anthropic2026sonnet5}, Claude Opus 5 \citep{anthropic2026opus5} and Kimi K3 \citep{moonshot2026kimik3} achieved the same scores. GPT-5.6 Luna \citep{openai2026gpt56} made one wrong selection, Qwen3.8 Flash \citep{qwen2026flash} one and Qwen3.8 Max 0902 \citep{alibaba2026qwen38max} five. All seven concerned the culture Choice.

Several configurations had fewer than fifty successful requests. GPT-5.6 Terra \citep{openai2026gpt56} and the two DeepSeek configurations \citep{deepseek2026v4pro0813, deepseek2026v41flash} answered every Choice in their successful requests correctly. Conditional scores in Appendix~\ref{app:observed} separate failed requests from wrong selections.

\subsection{Correct labels can conceal wrong quantities}
\label{sec:errors}

The culture case uses the experiments of \citet{luria1943mutations}. The ten samples in Experiment 11a come from one culture. The ten samples in Experiment 11 come from ten independent cultures. The two series therefore contain one and ten independent cultures, written $(1,10)$. The Choice asks whether the samples within each series come from one culture or from independent cultures.

Table~\ref{tab:culture-errors} shows the wrong selections across five repetitions of the culture Choice. Qwen Max selected three different wrong selections across its five repetitions. In all seven responses, the second Choice correctly identified colonies as clonal descendants. That selection contradicted the clause stating that each colony represented an independent mutation origin, so the final label remained \texttt{inconsistent} despite the wrong counts of independent cultures.

\begin{table*}[t]
\centering
\small
\begin{tabular*}{\textwidth}{@{\extracolsep{\fill}}lccccl@{}}
\toprule
& \multicolumn{2}{c}{Selection} & & & \\
\cmidrule(lr){2-3}
Configuration & Experiment 11a & Experiment 11
& \smash{\begin{tabular}[b]{@{}c@{}}Independent\\cultures\end{tabular}}
& \smash{\begin{tabular}[b]{@{}c@{}}Occurrences\\out of five\end{tabular}}
& Final label \\
\midrule
GPT-5.6 Luna     & Independent & One culture & $(10,1)$\phantom{0}            & 1 & \texttt{inconsistent} \\
\addlinespace
Qwen3.8 Flash    & Independent & One culture & $(10,1)$\phantom{0}            & 1 & \texttt{inconsistent} \\
\addlinespace
Qwen3.8 Max 0902 & One culture & One culture & \phantom{0}$(1,1)$\phantom{0}  & 2 & \texttt{inconsistent} \\
                 & Independent & One culture & $(10,1)$\phantom{0}            & 2 & \texttt{inconsistent} \\
                 & Independent & Independent & $(10,10)$                      & 1 & \texttt{inconsistent} \\
\bottomrule
\end{tabular*}
\caption{Wrong selections on the culture Choice. ``One culture'' means that the samples came from one culture, and ``Independent'' means that they came from independent cultures. Independent-culture counts are ordered as (11a, 11), with reference values $(1,10)$. Occurrences count how often each wrong selection appeared in five repetitions per configuration.}
\label{tab:culture-errors}
\end{table*}

Across the 588 successful requests, all final labels were correct, while downstream outputs were exact in 581. A label-only comparison would miss the seven requests whose downstream outputs were wrong.

\subsection{Reasoning tokens, latency and cost}

Reported reasoning tokens did not track correctness. Sol, Astra, Sonnet and Opus reported a median of zero reasoning tokens and answered every Choice correctly. On the culture Choice, DeepSeek V4 Pro used a number of reasoning tokens similar to Qwen Max's and answered all five repetitions correctly. Qwen Max erred in all five. The configurations that reported the most reasoning tokens were also the slowest, with median latencies of 5.4 to 14.2\,s.

Jev had the lowest median latency, 0.335\,s, with a p95 of 0.442\,s. Its mean cost was \$0.0595 per 1,000 requests. Among the configurations that had fifty successful requests and answered every Choice correctly, Sol had both the next lowest median, 1.385\,s, and the next lowest mean cost, \$3.0470 per 1,000 requests.

\section{Implications for Scientific Workflows}
\label{sec:discussion}

The comparison supports using Jev where evidence has been prepared, candidate relations are explicit and downstream operations are specified. Preserving the selections also makes it possible to inspect which selection caused a downstream discrepancy. Because four configurations reached complete correctness with a median of zero reasoning tokens, a workflow can reserve reasoning tokens for steps that cannot be reduced to a Choice, such as extracting evidence or proposing candidates.

The culture case shows that the gap reported by \citet{ho2025table} also affects quantities that a scientific workflow could reuse. Because a final label combines several clause judgments, one correctly contradicted clause can conceal an error elsewhere. Evaluations of components that select from fixed candidates should therefore keep both the reference answers and the reference downstream outputs.

\section{Conclusion}

We compared Jev with eleven other configurations on Choices with prepared evidence and candidates, using the same deterministic code for every configuration. Jev answered every Choice correctly. It was the fastest and least expensive configuration we measured. Four configurations reached complete correctness with a median of zero reasoning tokens. Separating selections, downstream outputs and final labels revealed errors hidden by the final label.

\section*{Limitations}

The comparison uses twenty model-resolved Choices in ten cases. The same curated English inputs recur across repetitions. Six configurations achieve complete correctness. All observed wrong selections concern one Choice. Evaluating transfer requires cases from experiments not represented in this collection and broader relation coverage.

The comparison cannot assess label discrimination, because every reference label is \texttt{inconsistent}. It cannot assess abstention, because no reference answer indicates insufficient evidence, conflicting evidence or an answer outside the listed candidates.

The evaluation begins with supplied evidence passages and structured records, so it does not cover extracting evidence or proposing candidates. The comparison uses one harness setup per model. The effect of individual harness settings, such as submitting both Choices in one request, is unmeasured. We did not vary reasoning effort within a model, so the comparison cannot show whether higher reasoning effort would help on harder cases.

Latency and cost depend on the serving provider, the time of the runs, reasoning effort and cache states. Charges for failed requests, data preparation costs and local computation costs are unmeasured.

\bibliography{references,case_sources}

\appendix
\input{appendix_material.tex}

\end{document}

%% file: appendix_material.tex
\section{Case Construction}
\label{app:cases}

The collection contains twenty case groups, forty scientific Choices and twenty-five claims. A case group shares evidence across its Choices and can support more than one claim. Ten case groups, comprising twenty scientific Choices, are code-resolved by deterministic rules developed on the prepared passages. The other ten each contain two model-resolved Choices and one claim. They are the ten cases of the model comparison. Tables~\ref{tab:code-cases} and~\ref{tab:model-cases} list the complete collection and the resulting computations. The collection cites twenty documents from eighteen sources. Two sections of one handbook count as one source, as do two documents on one cloud-seeding trial.

\begin{table*}[t]
\centering
\footnotesize
\setlength{\tabcolsep}{4pt}
\begin{tabular}{@{}L{0.06\textwidth}L{0.13\textwidth}L{0.28\textwidth}L{0.27\textwidth}L{0.18\textwidth}@{}}
\toprule
Group & Domain & Relation or unit & Downstream outputs & Source \\
\midrule
D01 & Concrete & Mixing operation and shared prism membership & One mixing operation, three prisms, six half-prisms & \citet{rezazadeh2026dataset} \\
D02 & Additive manufacturing & Complete build versus specimens within it & One build for three selected specimens & \citet{SHANBHAG2021107613} \\
D03 & Concrete curing & Bath-level control, run boundaries, specimen assignments & Curing-run count remains unknown & \citet{rezazadeh2026dataset} \\
C01 & Electroplating & Whole-plot unit for temperature/concentration, strip for current & One whole plot per setting, two strips per current level & \citet{nistUndatedNested} \\
H01 & Additive manufacturing & Column-level laser settings versus specimen-level heat treatment & Five selected columns, one heat-treated specimen & \citet{weaver2021demonstration} \\
C02 & Measurement systems & Wear-scar generation versus repeated scans & Two wear events distinguished by their cutter and wear-scar identifiers & \citet{wigginsUndatedMeasurement} \\
E001 & Semiconductor processing & Crossed implant/anneal memberships, wafer interaction unit & 64 setting combinations, one wafer each & \citet{nistUndatedNested} \\
E002 & Wind energy & Hourly controller switch and transition exclusion & Minute records do not add allocations, campaign count unknown & \citet{simley2021results} \\
E003 & Catalysis & Recovered catalyst carried between reaction cycles & Eight cycles in one documented reuse series & \citet{rafi2020copper} \\
E008 & Chemical process design & Equality of all eight coded factor settings & 20 runs, 18 distinct settings, two pure-error degrees of freedom & \citet{stallrich2026powerful}, \citet{nistUndatedFoldover} \\
\bottomrule
\end{tabular}
\caption{Relations and downstream outputs for the ten code-resolved case groups.}
\label{tab:code-cases}
\end{table*}

\begin{table*}[t]
\centering
\footnotesize
\setlength{\tabcolsep}{4pt}
\begin{tabular}{@{}L{0.06\textwidth}L{0.13\textwidth}L{0.28\textwidth}L{0.27\textwidth}L{0.18\textwidth}@{}}
\toprule
Case & Domain & Two Choices & Reference downstream outputs & Source \\
\midrule
E004 & Powder recycling & Mixed reuse histories\newline Partially known provenance & Exact parent-lot count unknown\newline Assignment incomplete & \citet{koushik2023effective} \\
E005 & Fatigue testing & Outcome-adaptive stress\newline Next test uses a new specimen & Next stress 295 or 355\,MPa after failure or runout & \citet{magazzeni2023bayesian} \\
E006 & Cosmology & Phase-reversed initial fields\newline Pair-average statistic & 100 realizations, 200 simulations, 100 ensemble values & \citet{villaescusa2018statistical} \\
E007 & Climate simulation & Two forcing protocols\newline Bundled implementation changes & 50 original and 50 SMBB members & \citet{ncarUndatedLens2} \\
E009 & Electrical metrology & Shared reference contribution\newline Bracketing average & Two laboratory results, three visits\newline Covariance retained & \citet{galliana2015extensive} \\
E010 & Cloud seeding & Complementary allocations\newline Allocation versus exposure & 118 joint assignments, 236 range-level records & \citet{rasmussen2018evaluation}\newline \citet{lawrence2014wyoming} \\
B001 & Developmental exposure & Injected F0 dams\newline Maternal, embryonic and germline exposure & 162 F3 pups\newline Zero new F3 injections & \citet{manikkam2012transgenerational} \\
B002 & Microbial genetics & One culture versus independent cultures\newline Clonal descendants & One versus ten independent cultures\newline Origins unknown & \citet{luria1943mutations} \\
B003 & RNA sequencing & RNA pooled before libraries\newline Aggregate expression profiles & Four libraries, 16 distinct contributors, 32 contributor appearances & \citet{rajkumar2015experimental} \\
B004 & Genome editing & Within-founder mosaicism\newline Inherited phenotype ambiguity & 23 founders, 57 reported alleles\newline Validated line count unknown & \citet{yen2014somatic} \\
\bottomrule
\end{tabular}
\caption{The ten cases in the model comparison. SMBB denotes smoothed biomass-burning forcing. Contributor appearances count each animal once for every pooled library to which it contributes.}
\label{tab:model-cases}
\end{table*}

\subsection{Evidence and reference construction}

Each input supplies a target, an explicit scope, selected evidence passages and structured records when the calculation requires them. The passages and records were prepared from articles, supplements, scientific project documentation, a methods handbook and an industrial case report.

For each of the twenty model-resolved Choices, we specified a reference answer and a rationale based on the supplied source evidence. One author reviewed these answers and their supporting rationales. Reference downstream outputs were derived from the reference answers and supplied records using the rules described for each case. Reference labels indicate whether the resulting quantities and relations support the constructed claims. Quantities not established by the evidence remained unknown.

Source explanations are part of the evidence, including explanations that directly establish a relation.

\subsection{Scientific interpretation of the model-resolved cases}

The descriptions below give, for each case, the two Choices, their reference answers and their alternatives. Alternatives are the other substantive candidates. Each Choice also lists candidates for missing evidence, for contradictory evidence and, where applicable, for an answer outside the listed candidates. Missing or ambiguous evidence maps to \texttt{not\_stated}. Unresolved incompatible evidence maps to \texttt{conflicting}. An established answer outside the listed candidates maps to \texttt{other}. E004 additionally distinguishes partially known provenance (candidate \texttt{partial\_identification}) from complete assignment and absent evidence.

\paragraph{E004: powder reuse histories}
From build 8 onward, recovered powder is topped up with previously reused powder, giving the feedstock mixed reuse histories and partial provenance. Alternatives are one common history, virgin replacement or separate unmixed histories. The second Choice asks whether each portion's original lot, mixture fraction and prior-build history are known. The reference is mixed histories with partial identification: a history label does not identify an original lot, so complete assignments and an exact parent-lot count remain unavailable.

\paragraph{E005: staircase fatigue testing}
The next stress depends on the preceding outcome and is applied to a new specimen. Alternatives are independently randomized, fixed or predetermined stresses, and continued loading of the same specimen or loading the whole foil before specimens are extracted. Starting at 325\,MPa with a 30\,MPa step gives 295\,MPa after failure and 355\,MPa after runout. The records give no outcome sequence, so the number of specimens tested at any one stress remains unknown.

\paragraph{E006: paired-fixed cosmological simulations}
The paired-fixed subset of the source's N1000 simulation set couples initial fields by phase reversal and contributes pair averages to the ensemble. Alternatives are independent draws, identical fields or output resampling, with individual simulations, pair differences or an overall mean as aggregation units. The counts in Table~\ref{tab:model-cases} describe this coupling and aggregation, not an effective sample size for every evolved observable.

\paragraph{E007: climate-ensemble protocol groups}
One Choice asks whether the ensemble splits into original and SMBB groups or follows a single protocol throughout. The other asks whether such a split reflects smoothing alone, smoothing bundled with other code changes or initialization only. The reference is the two-group split with documented emission and carbon-flux code corrections, so the comparison cannot isolate smoothing.

\paragraph{E009: correlated calibration uncertainty}
The results share reference-calibration uncertainty, and two secondary-laboratory visits form a bracketing average. Alternatives are independent realizations of that calibration or one result copied into both records. The visits could instead be two additional laboratories or a contrast across a deliberate adjustment. The squared uncertainty of the difference contains $-2u_B(\mathrm{std}_{\mathrm{DCV}})^2$, where $u_B(\mathrm{std}_{\mathrm{DCV}})$ is the standard uncertainty of the calibration of the shared 10\,V DC voltage standard. Effective sample size remains unknown.

\paragraph{E010: paired mountain-range allocation}
Complementary allocation makes each decision joint across the two ranges. The alternatives are independent assignments, seeding both or permanent assignment. Allocation establishes neither zero exposure nor confirmed exposure everywhere. The authors report 154 experimental units and exclude 27 for precipitation quality control and nine for generator performance. The remaining 118 joint assignments give 236 range-level records. Monthly generator-exclusion cells sum to eight, whereas the margins and prose report nine. The missing event remains unresolved.

\paragraph{B001: administration and developmental exposure}
Injection of pregnant F0 dams exposes the mother, embryo and fetal germline within the source's defined window. The administration alternatives include separate treatment of every generation, F3 animals alone and isolated germ cells. The exposure alternatives distinguish the full window from maternal-only exposure, mother plus embryo or exposure of every later descendant generation. The selected dioxin rows contain 162 F3 pups in 14 weaned litters, with zero newly injected F3 animals. Under the source's exposure-window definition, F3 is the first generation beyond that window. The number of distinct F0 ancestors remains unknown.

\paragraph{B002: cultures and mutation origins}
Section~\ref{sec:errors} gives the culture relation, and Table~\ref{tab:culture-errors} shows its three alternatives. The records cover the complete selected columns of source Tables~1 and~2, with 514 and 624 colonies. Several resistant colonies can descend from one earlier mutation, so the colony counts leave the number of mutation origins unknown. Alternatives are one origin per cell, origins created only at selection or repeated images of one colony.

\paragraph{B003: pooling and biological attribution}
RNA pooling before library preparation produces aggregate profiles. Alternatives are indexed-library pooling, numerical averaging or tissue pooling before extraction. For attribution, they are complete or partial individual profiles or a new animal. Four eight-animal libraries contain 16 distinct contributors and 32 contributor appearances, with zero separately observed contributor profiles. The individual-library arm remains in the evidence but outside the pooled-arm claim.

\begin{table*}[t]
\centering
\small
\begin{tabularx}{\textwidth}{@{}p{.06\textwidth}p{.44\textwidth}X@{}}
\toprule
Case & Output fields & Coverage of semantic distinctions \\
\midrule
E004 & Exact parent-lot count, availability of complete lot assignments. & The powder-history relation affects only a clause. \\
E005 & Next stress after initial failure, next stress after initial runout, observed replication at a fixed stress. & Specimen continuity is not used by the recorded stress calculation. \\
E006 & Paired realizations, simulations, values entering the ensemble. & The phase relation and the averaging operation are not recorded as output fields. \\
E007 & Original-protocol members, revised-protocol members, total members, whether the contrast changes smoothing only. & Different descriptions of a bundled change can share the same Boolean field. \\
E009 & Compared laboratory results, calibration visits, shared covariance term, effective sample size (unknown). & A bracketing average and an adjustment contrast can have the same recorded counts. \\
E010 & Original and retained experimental units, joint assignment events, range-level records, whether assignments to the two ranges are independent. & The exposure interpretation affects only a clause. \\
B001 & F3 pups, weaned litters, newly injected F3 animals, first generation beyond the original exposure window, distinct F0 ancestors. & Some alternative dosing descriptions imply the same newly injected F3 count. \\
B002 & Plating counts, independent-culture counts and colony totals by series, and exact mutation origins. & The colony--origin relation affects a clause, while the exact origin count stays unknown. \\
B003 & Selected libraries, distinct contributors overall and by genotype, contributor appearances, separately observed contributor profiles. & Pooling stage affects only a clause. \\
B004 & Deep-sequenced hybrid founders, reported mutant alleles, validated new germline lines. & The substantive mosaicism and marker alternatives affect only clauses. \\
\bottomrule
\end{tabularx}
\caption{Output fields used for downstream exactness. Because these fields capture only selected consequences of a Choice, semantic scoring covers distinctions that do not change them.}
\label{tab:output-fields}
\end{table*}

\paragraph{B004: mosaic founders and transmission}
One founder can carry several mutant alleles, so the 57 reported alleles do not count founders. The hybrid founders already carry an inherited albino allele, so albino offspring would not show that a new allele was transmitted. The number of validated new germline lines therefore remains unknown. Alternatives are a uniform allele pair, one founder per allele or no detected edit. For transmission, they are a unique new-allele marker, absent germ cells or verified allele-specific transmission. For founder 30, the author manuscript reports five alleles, whereas published Figure~3C lists six entries. Either count establishes more than two alleles, so the reference is mosaicism with founder 30's exact total unresolved.

\subsection{Deterministic derivation and label composition}
The shared code never substitutes zero for an unknown scientific quantity.

In the ten cases of the model comparison, selecting \texttt{not\_stated}, \texttt{conflicting} or \texttt{other} ends the derivation before the label rule in Section~\ref{sec:metrics} applies. The request then has no downstream outputs and no final label. The shared code records which candidate ended the derivation, with \texttt{conflicting} taking precedence over \texttt{other}, then \texttt{not\_stated}. Substantive candidates do not end the derivation. E004's partial provenance, for example, leaves one clause unresolved.

\section{Metric Definitions and Output Coverage}
\label{app:metrics}

Let $i\in\{1,\ldots,10\}$ index cases, $t\in\{1,\ldots,5\}$ repetitions and $k\in\{1,2\}$ Choices. Write $s_{itk}=1$ when the selection exactly matches its reference answer, and zero otherwise. Let $d_{it}$ indicate exact agreement on the case's output fields, and $\ell_{it}$ exact agreement on its final label. Failed requests set each indicator to zero. The metrics are
\begin{align}
 A_S &= \frac{1}{100}\sum_{i,t,k}s_{itk}, & A_D &= \frac{1}{50}\sum_{i,t}d_{it}, \\
 A_L &= \frac{1}{50}\sum_{i,t}\ell_{it}, & A_J &= \frac{1}{50}\sum_{i,t}d_{it}\prod_k s_{itk}.
\end{align}

Table~\ref{tab:main} reports the numerators of these four metrics. Downstream exactness requires agreement in value and type for every field in Table~\ref{tab:output-fields}. An omitted value differs from an explicitly unknown quantity, and a Boolean differs from a number.

E004, E007 and E010 store longer reference labels that also describe the individual clauses, such as \texttt{inconsistent\_in\_both\_clauses} for E007. Each of these claims has a contradicted clause, so we report its label as \texttt{inconsistent}.

\section{Correctness Conditional on a Successful Request}
\label{app:observed}

For a configuration with $n$ successful requests, conditional correctness uses the numerators in Table~\ref{tab:main} with denominators $2n$ for selections and $n$ for downstream, joint and label scores. For example, Qwen3.8 Flash scores $83/84$, $41/42$ and $42/42$ for semantic, downstream/joint and label correctness.

Across the 588 successful requests, 1,169 of 1,176 selections are correct.